\documentclass[10pt,twocolumn,letterpaper]{article}

\usepackage{cvpr}              

\usepackage{graphicx}
\usepackage{amsmath}
\usepackage{amssymb}
\usepackage{booktabs}

\usepackage[pagebackref,breaklinks,colorlinks]{hyperref}

\usepackage[capitalize]{cleveref}
\crefname{section}{Sec.}{Secs.}
\Crefname{section}{Section}{Sections}
\Crefname{table}{Table}{Tables}
\crefname{table}{Tab.}{Tabs.}

\def\cvprPaperID{3716} 
\def\confName{CVPR}
\def\confYear{2023}

\begin{document}

\title{BSNet: Lane Detection via Draw B-spline Curves Nearby}
\author{
Haoxin Chen\textsuperscript{1}  ~~Mengmeng Wang\textsuperscript{1} ~~ Yong Liu\textsuperscript{1*} \\
\textsuperscript{1}  APRIL Lab, Zhejiang University, China  
}

\maketitle
\let\thefootnote\relax\footnotetext{*Corresponding Author.}
\begin{abstract}

Curve-based methods are one of the classic lane detection methods. They learn the holistic representation of lane lines, which is intuitive and concise. However, their performance lags behind the recent state-of-the-art methods due to the limitation of their lane representation and optimization. In this paper, we revisit the curve-based lane detection methods from the perspectives of the lane representations' globality and locality. The globality of lane representation is the ability to complete invisible parts of lanes with visible parts. The locality of lane representation is the ability to modify lanes locally which can simplify parameter optimization. Specifically, we first propose to exploit the b-spline curve to fit lane lines since it meets the locality and globality. Second, we design a simple yet efficient network BSNet to ensure the acquisition of global and local features. Third, we propose a new curve distance to make the lane detection optimization objective more reasonable and alleviate ill-conditioned problems. The proposed methods achieve state-of-the-art performance on the Tusimple, CULane, and LLAMAS datasets, which dramatically improved the accuracy of curve-based methods in the lane detection task while running far beyond real-time (197FPS). 
   
\end{abstract}

\begin{figure}
  \centering
  \begin{subfigure}{0.49\linewidth}
    \includegraphics[width=1\linewidth]{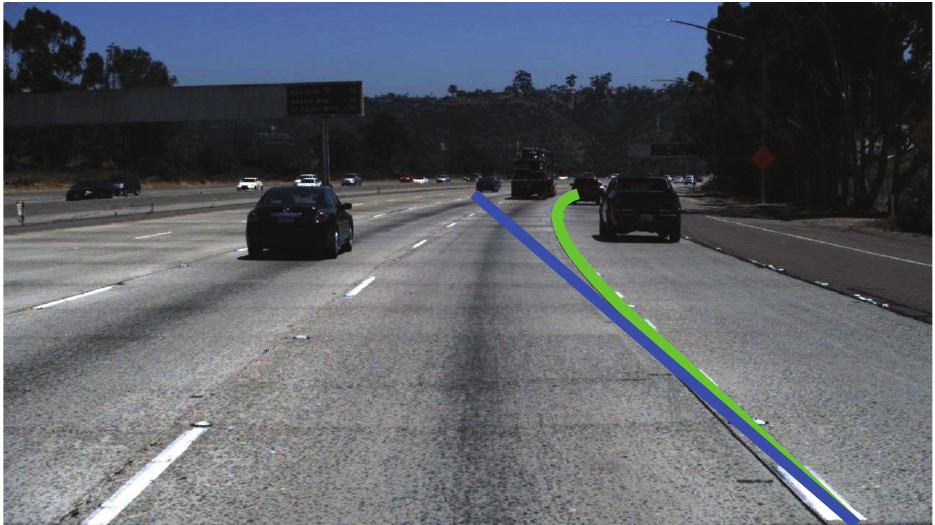}
    \caption{}
    \label{fig:pred_and_groundtruth}
  \end{subfigure}
  \hfill
  \begin{subfigure}{0.49\linewidth}
	\includegraphics[width=1\linewidth]{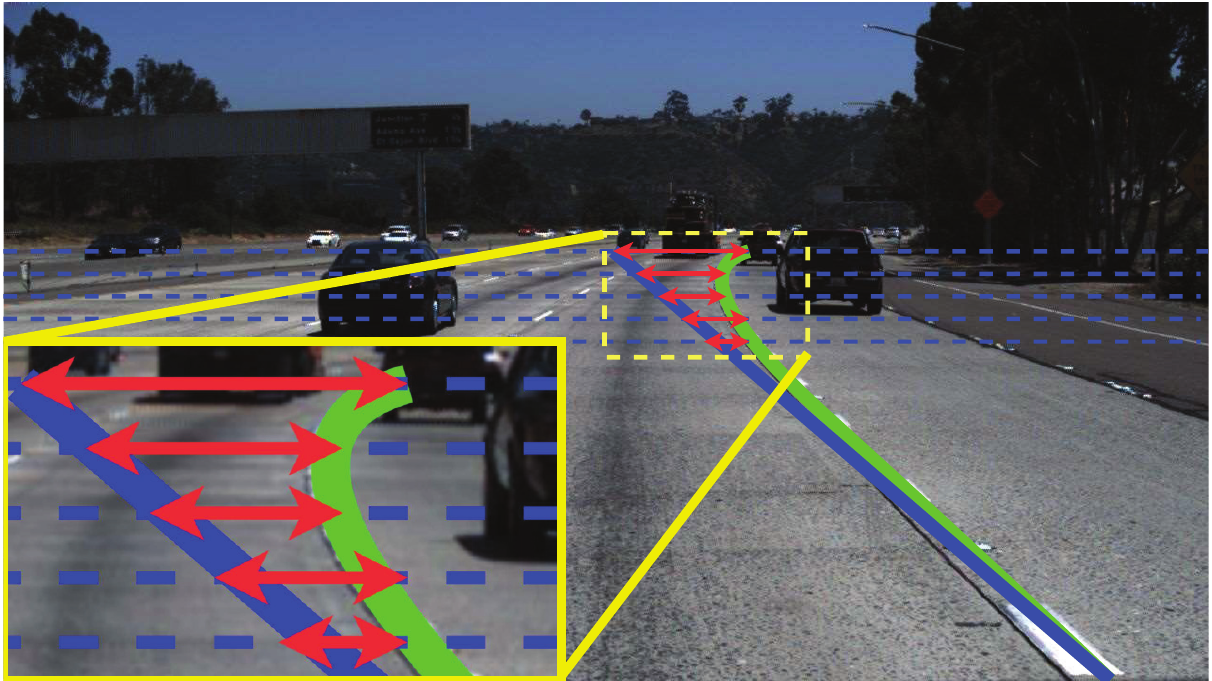}
	  \vfill
	\caption{}
	\label{fig:polynomial}
\end{subfigure}
  \begin{subfigure}{0.49\linewidth}
	\includegraphics[width=1\linewidth]{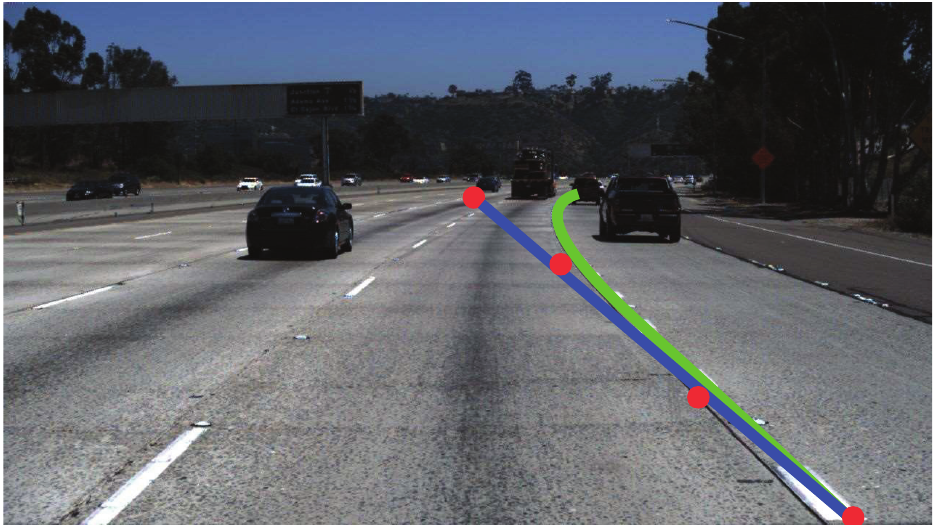}
	
	\caption{}
	\label{fig:bezier}
\end{subfigure}
  \hfill
  \begin{subfigure}{0.49\linewidth}
	\includegraphics[width=1\linewidth]{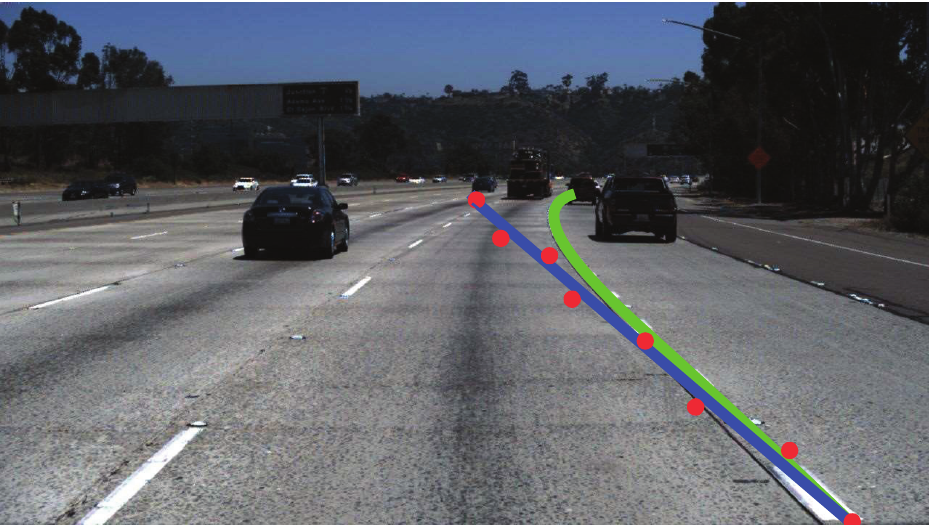}
	
	\caption{}
	\label{fig:bspline}
\end{subfigure}
  \caption{When facing the same training situation using different lane representations, what will happen after backpropagation? (a) The green curve is the ground truth of lane, and the blue curve is the predicted lane when training. (b) Lane representation is polynomial curves, and offsets are in red. The part of curves with big offsets is magnified. (c) Lane representation is bézier curves. (d) Lane representation is b-spline curves.}
  
\end{figure}

\section{Introduction}
\label{sec:intro}

Lane detection task is an important component in autonomous driving systems. It is helpful for vehicles to determine their current positions, which in turn affects the subsequent decision-making planning. There are many challenges for this task, such as lane lines occluded by vehicles, faded lane lines, dashed lane lines, and adverse illumination or weather conditions.

In recent years, deep-learning based lane detection methods have shown their promising prospects. Especially, we divide them into five categories: segmentation-based, keypoint-based, line-anchor-based, row-anchor-based, and curve-based methods. Among them, the parametric curve is an intuitive and concise lane representation, but the performance of the curve-based methods is inferior to the state-of-the-art methods. To excavate the underlying causes, we provide an in-depth analysis based on the existing lane detection methods and observe two significant characteristics of lane representation: globality and locality. Formally, the globality of lane representation is the ability to complete invisible parts of lanes with visible parts. The locality means the ability to modify lanes locally.

More specifically, the globality of lane representation is very beneficial in predicting holistic lanes since it makes completing invisible parts efficient. Line-anchor-based methods\cite{li2019line,tabelini2021keep,su2021structure,zheng2022clrnet} predict small offsets, then the globality obtained from line anchor priors doesn't be broken. Curve-based methods' lane representations are holistic curves and have natural globality. The segmentation-based methods\cite{pan2018spatial,zheng2021resa,neven2018towards,abualsaud2021laneaf,xu2020curvelane}, keypoint-based methods\cite{ko2021key,qu2021focus,wang2022keypoint}, and row-anchor-based methods\cite{qin2020ultra,liu2021condlanenet,qin2022ultra} lack lane representation globality. They usually use low-efficient structures to make up for the lack of lane representation globality and get holistic lanes, such as message-passing mechanisms and post-processing.

The locality of lane representation makes the parameter optimization more efficient. We show the importance of locality with an example.  Two curves are shown in \cref{fig:pred_and_groundtruth}, the blue curve and green curve are predicted lane and ground truth. The lower parts of the two curves are almost aligned, higher parts have big offsets. Past curve-based methods use polynomial or bézier curves as lane representations, which lack the locality. Each parameter of the curve changes will cause all points on the curve to move. If we use polynomial or bézier curves to represent curves as shown in \cref{fig:polynomial,fig:bezier}, and calculate the loss between ground truth and predicted lane, the higher parts will generate a big loss, the lower parts' loss is zero. After backpropagation, the new parameters will let higher parts closer while the lower parts are less considered, then the distance between lower parts will be farther. There is a kind of competition between different parts of the curve, making the network hard to optimize, which cannot fit lanes well and has lower accuracy. Suppose there are much more straight lanes in training datasets than curvy lanes, the network will intend to overfit straight lanes due to the competition between the curve's different parts, which is one reason curve-based methods perform poorly in curvy lane scenes. Other methods besides curve-based methods predict each part of lanes independently thus no competition between different parts of lane line and fit lane flexibly.

The b-spline curve meets globality and locality meanwhile. Points on the k-th degree b-spline curve are only judged by k+1 control points nearby to guarantee the locality. Using b-spline curves to fit curves as shown in \cref{fig:bspline}, we only need to move higher control points, then the higher parts of curves will be closer and the parameters will be optimized efficiently. In addition, b-spline curves can increase the number of control points when facing lanes with complex topology, and the curves' power does not change. If we increase the number of bézier curves' control points, the curve's power will increase, which is harmful to parameter optimization. We observed some ill-conditioned problems due to the imperfect distance calculation method between curves when training curve-based networks, which caused the optimization objective to not match the true objective and networks to be optimized inefficiently. In particular, we propose a new curve distance to solve these problems. We design an efficient network BSNet inspired by segmentation networks, which 
could satisfy the requirements of the global and local features efficiently. Our methods achieve state-of-the-art results on Tusimple\cite{tusimple}, CULane\cite{pan2018spatial}, and LLAMAS\cite{llamas2019} datasets. Meanwhile, our network has a very fast speed of 197 FPS. The main contributions can be summarized as follows:
 \begin{itemize}
 		\item[$\bullet$] We propose a novel b-spline curve-based deep lane detector, which meets globality and locality of lane representation meanwhile and thus has a stronger fitting ability.   
		\item[$\bullet$] We propose a novel curve distance calculation method, alleviating ill-conditioned problems due to imperfect distance calculation methods, and the optimization of networks is more efficient.
		\item[$\bullet$] We propose a simple yet efficient network structure, and it meets the demand for global and local features at the same time.
		\item[$\bullet$]   Our methods achieve state-of-the-art results on Tusimple, CULane, and LLAMAS datasets. And our network has a small detection head. Even our ResNet-34 version model is faster than  the ResNet-18 version CLRNet\cite{zheng2022clrnet}. It's more friendly to be combined into multitasking networks. 
\end{itemize}


  \begin{figure*}[tp]
	\includegraphics[width=0.9\linewidth,scale=1.00]{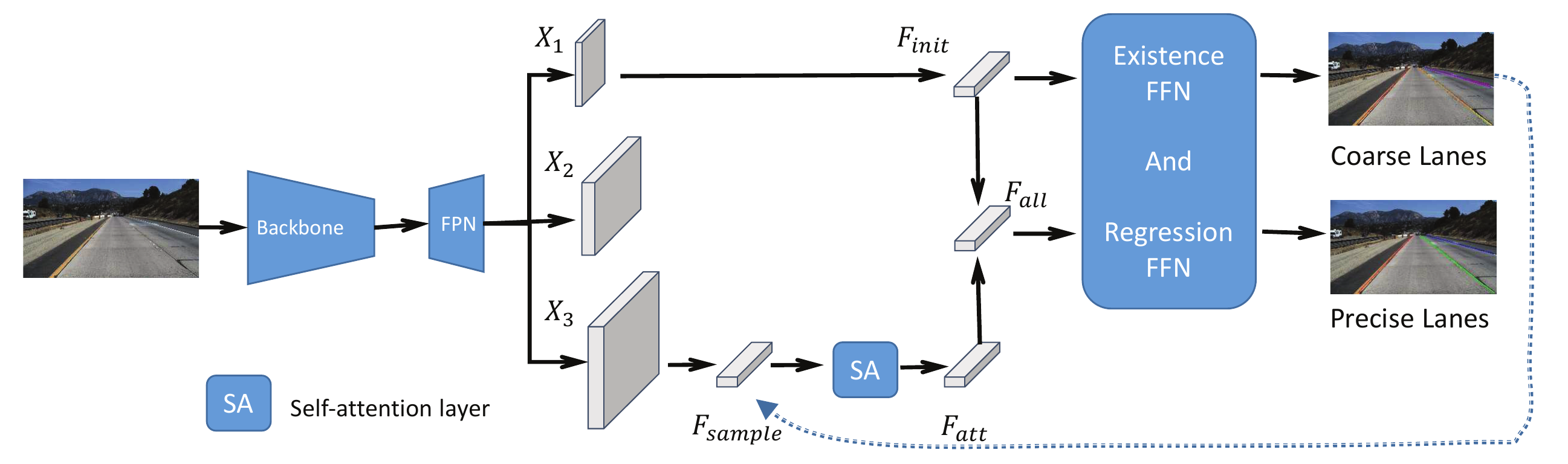}
	\centering
	\caption{ The structure of BSNet. An image is fed into ResNet and FPN, and the smallest resolution feature $ X_1 $ is utilized to generate lane proposals' init feature $  F_{init} $. The sample points feature $ F_{sample} $  of lanes is obtained from the largest resolution feature $ X_3 $ and is processed with a self-attention layer to get $ F_{att} $. The final lanes' feature $  F_{all} $ is the sum of $  F_{init} $, $ F_{sample} $ and $ F_{att} $.  }
	\label{fig:network}
\end{figure*}

\section{Related work}
\subsection{Segmentation-based methods}
These methods treat lane detection as a pixel-wise classification task and get holistic lanes through global features and post-processing. SCNN\cite{pan2018spatial} obtains global features from a message-passing mechanism that helps each pixel to get long-range information. Therefore, the invisible parts of lanes can be completed. But its latency is very high due to the high computation cost. RESA\cite{zheng2021resa} optimizes the efficiency of the message-passing mechanism, and the speed and accuracy are improved. LaneNet\cite{neven2018towards} uses clustering to solve the problem of detecting a dynamic number of lanes. LaneAF\cite{abualsaud2021laneaf} proposes using affinity fields to decode foreground areas to different lane instances. Segmentation-based methods have medium accuracy, but their computation cost is high, and they need post-processing to get lane instances.

\subsection{Keypoint-based methods}

PINet\cite{ko2021key} utilizes heatmap to locate approximate locations of lanes' keypoints, then utilizes offsets to refine to accurate locations and obtains lane instances after post-processing. FOLOLane\cite{qu2021focus} is a bottom-to-up lane modeling method. It first detects local keypoints, then uses offsets between adjacent keypoints to decode holistic lane curves. GANet\cite{wang2022keypoint} proposes using offsets between keypoints and lanes' start points to constrain the lane representation globality and proposes LFA to get features of holistic lanes. Keypoint-based methods get lane instances through post-processing, which is low-efficient. And the resolution of the predicted heatmap can't be too small, or the accuracy will decrease, which means the network is hard to be very lightweight.

\subsection{Line-anchor-based methods}

These methods get lane predictions by predicting offsets to line anchor priors. Line-CNN\cite{li2019line} is the pioneer of line anchor representation and uses features of pixels where line anchors' start points are located to predict offsets. LaneATT\cite{tabelini2021keep} proposes anchor-based feature pooling to utilize features efficiently and utilizes an attention mechanism to strengthen global features so that it can use a lightweight backbone. Its speed is much faster than Line-CNN. CLRNet\cite{zheng2022clrnet} proposes a refinement mechanism to utilize low-level and high-level features and proposes ROIGather to gather global context. Line-anchor-based methods have high accuracy, but they can't predict nearly horizon lanes due to the inherent shortcoming of line anchors. Their performance relies on the position of line anchor priors.

\subsection{Row-anchor-based methods}
UFLD\cite{qin2020ultra} first proposes using row anchors to detect lanes. The lane detection task is converted to a row-wise classification task. They obtain global features from a fully connected network and have fast speed, but accuracy is lower than other methods. CondLaneNet\cite{liu2021condlanenet} proposes using conditional convolution to improve lanes' features and utilizes RIM to overcome the problems when facing lanes with complex topology. UFLDv2\cite{qin2022ultra} proposes a hybrid anchor and uses ordinal classification to improve UFLD's performance. UFLDv2 maintains high speed and improves accuracy, but the accuracy is still lower than state-of-the-art methods (i.e., CLRNet\cite{zheng2022clrnet}, GANet\cite{wang2022keypoint}).

\subsection{Curve-based methods}

PolyLaneNet\cite{tabelini2021polylanenet} proposes using polynomial curves as lane representation. LSTR\cite{liu2021end} utilizes DETR-like\cite{carion2020end} structure and geometry constraints of lanes to improve detection performance. Laneformer\cite{han2022laneformer} uses a transformer encoder and tries to exploit the relationship between occluded parts of lanes and vehicles. Under the guidance of the vehicle's position information, the occluded lanes are completed. BézierLaneNet\cite{feng2022rethinking} proposes using bézier curves as lane representation and a feature flip fusion module to exploit the symmetry property of lanes. Its accuracy is higher than methods based on polynomial curve representation but still lags behind the state-of-the-art methods. Curves are intuitive and concise representations of lanes, curve-based methods have fast speed, but their accuracy is not satisfied.

\section{Method}

\subsection{The Lane Representation }
We choose the clamped quasi-uniform b-spline curve as lane representation. It has the locality required for lane line fitting and can flexibly fit lane lines. Bézier curve's start point and end point are control points, so the starting and ending positions of the lane lines can be more intuitively reflected from the parameters. The clamped quasi-uniform b-spline curve have similar characteristic. The clamped quasi-uniform b-spline curve is defined by $ n + 1 $ control points \(P_0, P_1, ..., P_n\), and $ m+1 $ knots \(u_0=u_1=...=u_p=0, u_{p+1}, u_{p+2}, ..., u_{m-p-1}, u_{m-p}=u_{m-p+1}=...= u_m=\)1. The formulation of the b-spline curve of degree \(p\) is shown in  \cref{eq:bspline}, which is similar to the formulation of the bézier curve. The position of each point on the curve is calculated by the weighted sum of the control points:

\begin{equation}
C(u) = \sum\limits^n_{i=0} N_{i,p}(u)P_i
  \label{eq:bspline}
\end{equation}
where \(P_i\) is the \(i-th\) control point, \(N_{i,p}(u)\) is the b-spline curve basis function of degree \(p\):
\begin{equation}
N_{i, 0}(u)=\left\{\begin{array}{ll}
1 & \text { if } u_{i} \leq u<u_{i+1} \\
0 & \text { otherwise }
\end{array}\right.
  \label{eq:basis1}
\end{equation}
\begin{equation}
\begin{aligned}
N_{i, p}(u) =&\frac{u-u_{i}}{u_{i+p}-u_{i}} N_{i, p-1}(u)\\
&+\frac{u_{i+p+1}-u}{u_{i+p+1}-u_{i+1}} N_{i+1, p-1}(u)
\end{aligned}
  \label{eq:basis2}
\end{equation}where \(u_i\) is the \(i-th\) knot, and the half-open interval \([u_i, u_{i+1})\) is the \( i-th\) knot span. For clamped quasi-uniform b-spline curve, in addition to the first knot and the last, other knots should be uniformly distributed. 

To generate b-spline ground truth, similar to the method in BézierLaneNet\cite{feng2022rethinking}, we fit the labeled points of the lanes in datasets to b-spline curves. \(N_{dis}\) is the number of sample points on each curve. Sample points at equal intervals of \(u\) are utilized as the targets for loss calculation.

\subsection{Proposals' features init module}
\begin{figure}[tp]
    \includegraphics[width=0.8\linewidth]{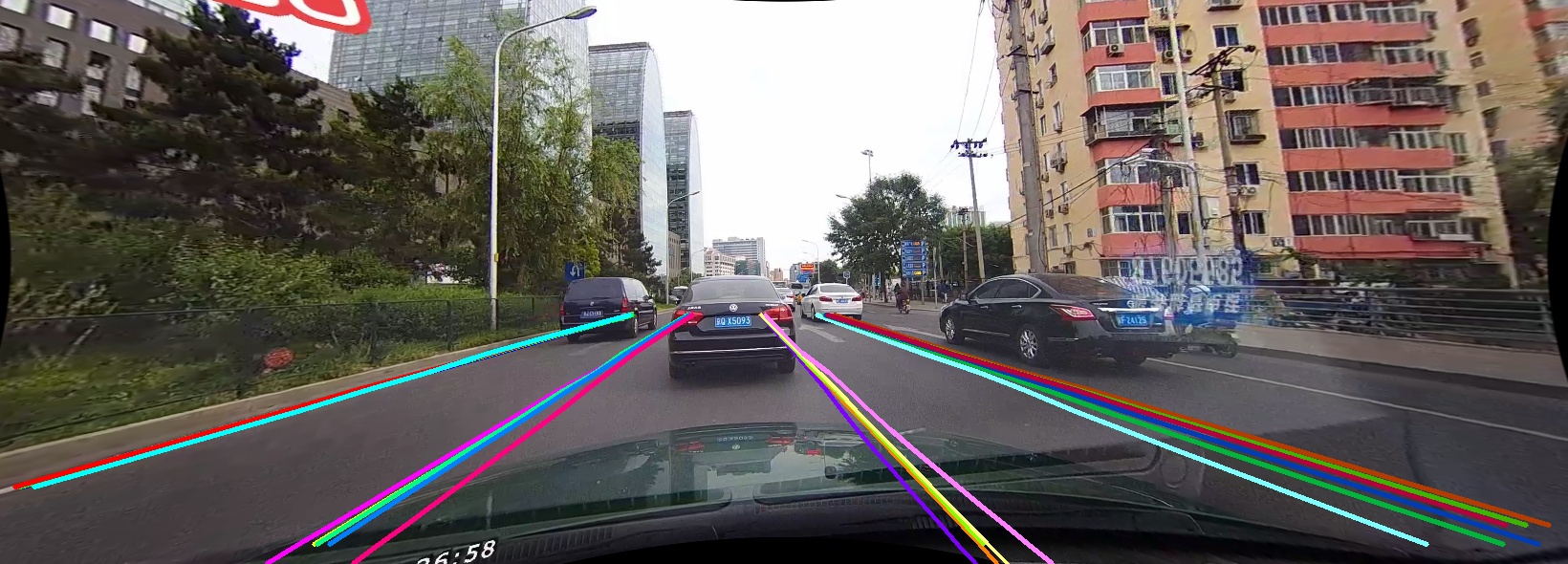}
     \centering
	 \caption{An example of initial lane proposals. We show lane proposals in which existence confidence is over 0.1 here.}
	 \label{fig:initial_proposals}
  \end{figure}

This module outputs lane proposals' features corresponding to each image. If we use segmentation-based methods without post-processing, although there will be noisy points and occluded parts of lanes ignored, lanes' coarse outlines are detected. This means each lane's information can be kept in features of all pixels in one channel. Inspired by this, we designed a proposals' features init module. Specifically,
our backbone is ResNet\cite{he2016deep}, and neck is FPN\cite{lin2017feature}, features output by FPN in order of increasing resolution are \(X_1,X_2,X_3\). The shape of \(X_1\) is \( 512 \times \frac{H }{32} \times  \frac{W }{32} \), then it will be reshaped to feature \(X_{1\_reshape}\)  (\(512 \times \frac{HW}{1024}\)), we feed \(X_{1\_reshape}\)   into a feed-forward network (FFN) to get spatial information of every channel, the output is \(X_{1\_spatial}\)  (\(512 \times C\)). At last, we use a \(1\times 1\) 1D convolution to get \(X_{init}\)  (\(N_p\times C\)), which is the init features of lane proposals, \(N_p\) is the number of proposals,  \(C\) is the feature dimension of each proposal. 

\(X_{init}\)  is fed into the class prediction module and regression module to get lanes' existence confidence and control points' coordinates. The class prediction module and regression module consist of  FFN layers. The prediction results are a set of curves whose shape is similar to lanes, and we call them lane proposals or coarse lanes, one of the results is shown in \cref{fig:initial_proposals}.

\subsection{Local features and attention mechanism}

The features we obtain from the proposals' features init module are global, and we need to add more local information.
Similar to feature pooling operations in\cite{tabelini2021keep,zheng2022clrnet,girshick2015fast}, we pool features on predicted lane proposals' sample points as local feature \(F_{sample}\), and the number of sample points is $N_{fea}$. Then \(F_{sample}\)  is fed into a standard self-attention layer to get feature \( F_{att}\) which keeps information of the relationship between lane proposals. We add \(F_{init},F_{sample},F_{att}\) together to get lane proposals' final feature \(F_{all}\):
\begin{equation}
F_{all} = (F_{init}+F_{sample}+F_{att})
    \label{eq:feature_all}
\end{equation}
The \(F_{all}\) will be fed into the class prediction module and regression module too, the prediction results are the precise lane locations of our network, and fast non-maximum suppression (Fast NMS)\cite{bolya2019yolact} is utilized to obtain final results. Fast NMS achieves fast speed and doesn't rely on CUDA extension, which is more friendly to deploy.

\begin{figure}
	\centering
	\begin{subfigure}{0.49\linewidth}
		\includegraphics[width=1\linewidth]{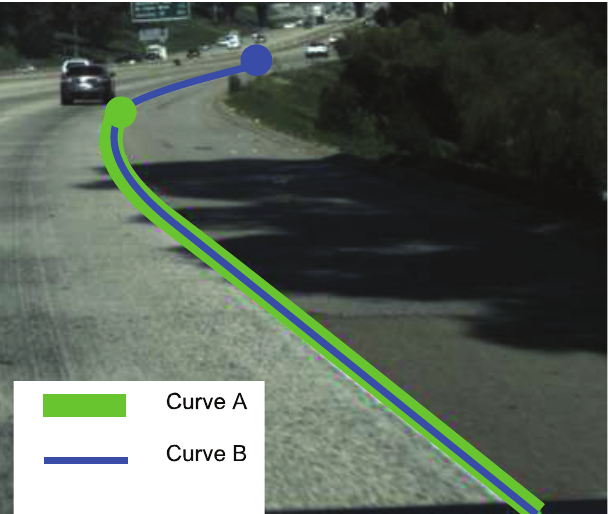}
		\caption{}
		\label{fig:too_far_img}
	\end{subfigure}
	\hfill
	\begin{subfigure}{0.49\linewidth}
		\includegraphics[width=1\linewidth]{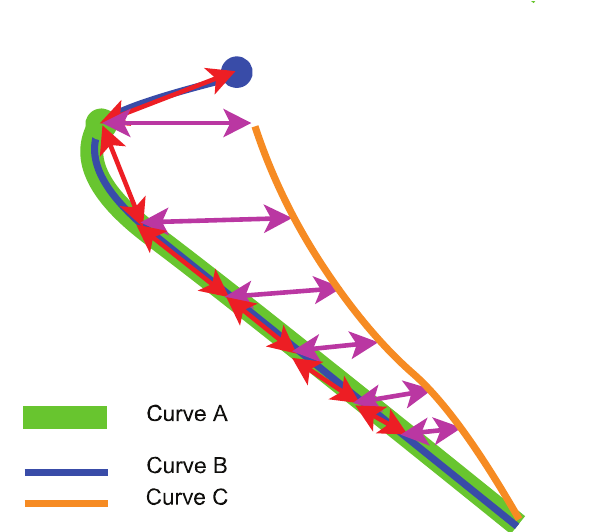}
		\caption{}
		\label{fig:too_far_equivalent}
		
	\end{subfigure}
	\caption{Bézier curves A, B, and C are ground truth, predicted lane, and a distance equivalent curve of the predicted lane. (a) The position of ground truth and predicted lane in the real world is shown, blue point and green point are endpoints of curves. (b) Red arrows represent the distance between sampling points that have the same $ u $ when calculating the distance between curve B and curve A. In order to show the distance intuitively, we draw an equivalent curve C of curve B. Distance from curve B or curve C to curve A is equal. }

	\label{fig:ill_conditon}
\end{figure}

\subsection{Distance between curves}
The previous distance calculation methods\cite{tabelini2021polylanenet,feng2022rethinking} between curves don't fully align with our optimization objective. For instance, PolyLaneNet\cite{tabelini2021polylanenet} calculates the distance between sample points on the uniformly sampled row lines as the distance between curves. This way is friendly to vertical lanes but will lead to sparse supervision for horizontal lanes because horizontal lanes only occupy a few row lines, and it can't measure the distance between horizontal curves very well. UFLDv2\cite{qin2022ultra} mentions this kind of magnified error too, and shows that the error band of this kind of calculation method is multiplied by a factor of \(\frac{1}{sin\theta}\). When angle \(\theta\) between lane and sampled row lines is small, the magnifying factor will go to infinity.
Recently, BézierLaneNet\cite{feng2022rethinking} proposes to calculate the distance of curves as the average of distances between sampling points with the same $ u $, and the distance between bézier curves A and B is  $ D_{AB} $. We show three bézier curves in \cref{fig:ill_conditon} to illustrate the problem of using their calculation method in our framework. 
Specifically, curve A is the labeled ground truth, curve B is the predicted lane, and curve C is the equivalent distance curve of curve B, which means $ D_{AB}=D_{AC}$. The predicted curve B contains farther parts of the lane out of the labeled lane, which caused $ D_{AB} $ to be big because the sampling points which have the same $ u $ are far from each other, although holistic curves are close to the real-world lane. Due to curve A and curve B are coincident, we utilize the distance equivalent curve to intuitively show the distance between curve B and curve A, curve C and curve A have big offsets, and $ D_{AC}=D_{AB} $, this means that if we calculate the distance to curve A, curves between curve C and curve A can get smaller value than curve B, but curve B is more similar to real-world lane. This reflects that the distance calculation method of BézierLaneNet is in conflict with our optimization objective sometimes, causing an ill-conditioned problem. If we label the picture in \cref{fig:too_far_img}, it is reasonable to label the lane as either curve A or curve B, and both cases are possible since there is no rule for where the endpoint should be labeled. So this distance calculation method is imperfect, what we really need is a curve that is closer to the real-world lane. How can we solve this ill-conditioned problem?

We propose a new curve distance to solve this problem. The objective of lane detection is to make prediction curves as close as possible to ground truth, which means each point \(P\) on curve \(B\) should be as close as possible to the point on curve \(A\) which is closest to point \(P\). We define \(D_{P\rightarrow A}\)  as the distance from point P to curve A. It is the distance from point \(P\) to the closest point on curve \(A\) too. \(D_{P\rightarrow A}\)  is formulated as follows:
\begin{equation}
\begin{aligned}
D_{P\rightarrow A}= min(L_2(P,P_{0}),&L_2(P,P_{1}),\cdots,\\
&L_2(P,P_{N_A-1})) 
\end{aligned}
\label{eq:point_to_curve}
\end{equation}
where \(N_A\) is number of points on curve A, \( P_0,P_{1},\cdots,P_{N_A-1}\)  are points on curve A. Then the distance from curve A to curve B \(D_{A\rightarrow B}\) is calculated as follows:
\begin{equation}
	D_{A\rightarrow B} = \frac{1}{N_A}\sum_{i=0}^{N_A} D_{P_i\rightarrow B} 
	\label{eq:curve_distance}
\end{equation}
In \cref{fig:ill_conditon}, if we utilize \cref{eq:curve_distance} as the distance between curves,   \(D_{B\rightarrow A}\)  is smaller than \(D_{C\rightarrow A}\), so the distance between curve A and curve B is smaller than the distance between curve A and curve C, then the ill-conditioned problem is alleviated. But if curve \(B\) degenerates to a point on curve \(A\), $D_{B\rightarrow A }$ will become zero. In order to solve the degradation problem, we set \(D_{B\rightarrow A}+D_{A\rightarrow B}\)  as distance between curve \(A\) and \(B\):

\begin{equation}
D_{AB}=D_{B\rightarrow A}+D_{A\rightarrow B}
    \label{eq:curve_distance_double}
\end{equation}
Due to curves being continuous, there are infinity points on curves, and the \cref{eq:point_to_curve,eq:curve_distance} can't be calculated numerically. Firstly, we calculate \(N_{dis}\) sample points' position on each curve, and then we connect the adjacent sample points into line segments. If we want to calculate \(D_{P\rightarrow A}\), it can be approximated as calculating the distance \(D_{P\rightarrow segment}\)  from point \(P\) to the closest line segment on \(A\) to \(P\). We define the closest two points on \(A\) to \(P\) are \(P1, P_2\). They can consist triangle with P, and we regard point P as the upper point, \(PP_1,PP_2\) are triangle waists. If base angles are acute, \(D_{P\rightarrow segment}\)   is equal to triangle high, otherwise equal to \(min(PP_1,PP_2)\).

 When the distance is utilized for loss, We  use IoU-like format to normalize \(D_{P\rightarrow A}\) to $ D_{P\rightarrow A}^N $ as follows:

\begin{equation}
	D_{P\rightarrow A}^N=\frac{2r-D_{P\rightarrow A}}{D_{P\rightarrow A}+2r}
	\label{eq:IoU}
\end{equation}
where $ r $ is the extended radius of lane line.  
  \(D_{A\rightarrow B}^N\) can be approximated as average of  \(D_{P_i\rightarrow B}^N\ (i= 0,1,\cdots,N_{dis})\), in here,  \(P_i\) is sample point on \(A\). 
 If lane's ground truth \(C_{gt}\) and predicted lane \(C_{pred}\)  are given,  the regression loss \(L_{reg} \) can be calculated similar with \(D_{AB}\):
\begin{equation}
L_{reg}=1- \frac{1}{2}(D_{C_{gt}\rightarrow C_{pred}}^N+D_{C_{pred}\rightarrow C_{gt}}^N)
    \label{eq:loss_reg}
\end{equation}

\subsection{Label assignment}
 We assign labels by the position of the start points of the lanes. \(N_{p}\)  is the number of proposals, we set \(N_{p}\)
reference points on the left, bottom, and right borders of image,  the left or right border have \(\frac{N_{p}}{4}\) reference points, the bottom border has  \(\frac{N_{p}}{2}\) reference points.  Reference points are uniformly distributed on each border, and every reference point corresponds to a lane proposal.  If we want to assign proposals to a ground truth lane, we calculate the distance between the lane's start point and reference points, then we select the top-k reference points closest to the start point, and the corresponding proposals' category will be positive, the regression loss between these proposals' predicted lanes and ground truth will be calculated.

It is worth noticing that our reference points are only auxiliary tools for label assignments. They have nothing to do with the start points of line anchors used in line-anchor-based methods. The lane proposals' init features and coarse lanes are predicted corresponding to each image, rather than fixed like line anchors, so our method doesn't rely on any prior information.

\begin{figure}
  \centering
  \begin{subfigure}{0.45\linewidth}
    \includegraphics[width=1\linewidth]{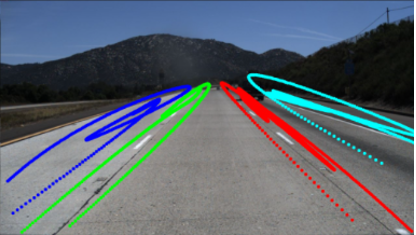}
    \caption{}
    \label{fig:graffiti}

  \end{subfigure}
  \hfill
  \begin{subfigure}{0.45\linewidth}
    \includegraphics[width=1\linewidth]{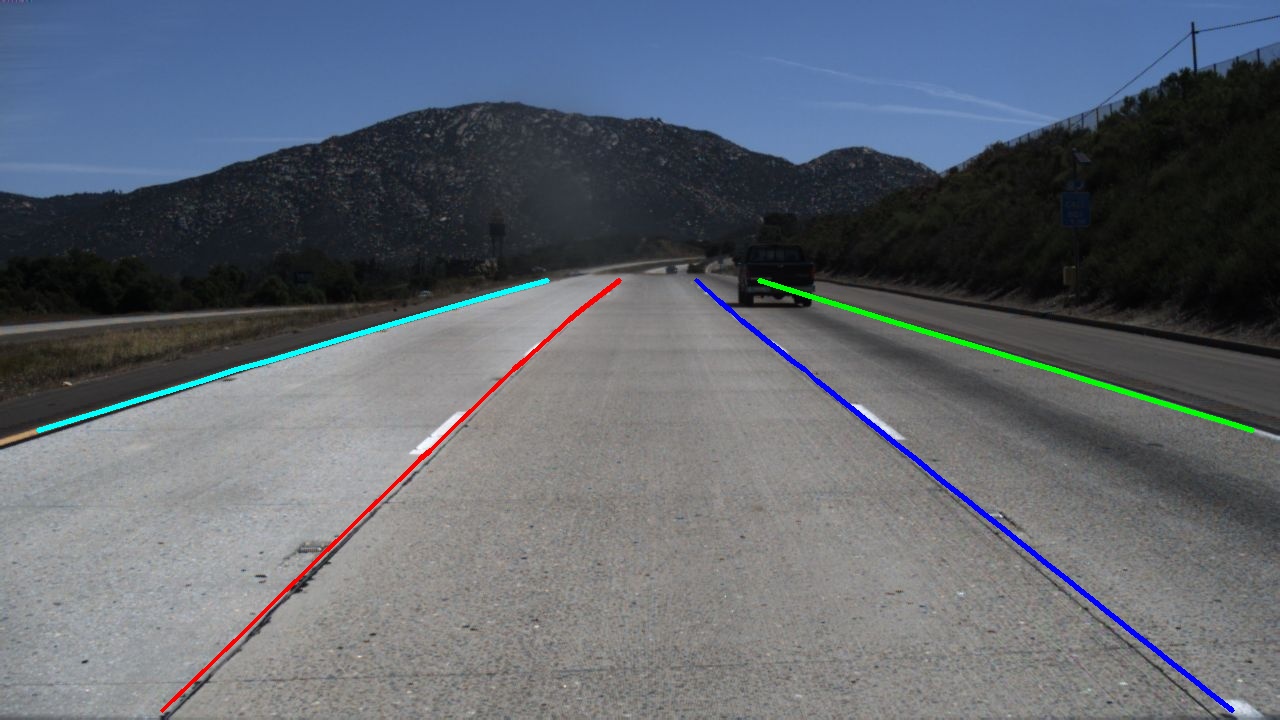}
    \caption{}
    \label{fig:perfect}
  \end{subfigure}
  \caption{(a) The predicted chaotic graffiti lanes. (b) The predicted lanes with simple topology.}
  
\end{figure}

\subsection{Overall loss}

  The regression loss $ L_{reg} $ can make predicted lanes close to ground truth, but there are some other problems in fact. If we only use regression loss \(L_{reg}\)  and classification loss \(L_{cls}\) in training, the predicted lanes will be chaotic graffiti, and they are closer and closer to ground truth along training, we show an example of predicted lanes in early training term in \cref{fig:graffiti}. We want to get predicted lanes in the form of curves with relative simple topology rather than chaotic graffiti. We add a length loss \(L_{length}\)  to solve this problem because although the result in the form of graffiti is close to the ground truth, it will lead to the curve's length being much longer than the ground truth. The formulation of  \(L_{length}\)  is:  
\begin{equation}
L_{length}= \frac{abs(l_{gt}-l_{pred})}{l_{gt}}
    \label{eq:loss_length}
\end{equation} where \(l_{gt},l_{pred}\) are lengths of predicted curve and ground truth lane. They can be calculated by summing the length of line segments which connected sample points. By now, a perfect prediction result with a relative simple topology can be obtained, as shown in \cref{fig:perfect}. In addition, in order to make the optimizing process more stable, we add the start point loss \(L_{start}\):

\begin{equation}
	L_{start}= MSE(S_{gt},S_{pred})
    \label{eq:loss_start}
\end{equation}
where \(S_{gt},S_{pred}\) are start points of predicted curve and ground truth lane, \( MSE\) is the mean square error loss. \(L_{start}\) prevents the control points at start and end points from being interchanged during the optimizing process, thus avoiding the instability of the training process. Our \(L_{cls}\) is the focal loss\cite{lin2017focal} between predictions and ground truth. The overall loss function is shown following:
\begin{equation}
	L_{total}= \lambda_1  L_{reg}+\lambda_2  L_{length}+\lambda_3  L_{start}+\lambda_4  L_{cls}
    \label{eq:loss_total}
\end{equation}where \(\lambda_1,\lambda_2 ,\lambda_3 ,\lambda_4 \) are constants.

\section{Experiments }

\subsection{Datasets}

We conduct experiments on three widely used benchmarks:  CULane\cite{pan2018spatial}, Tusimple\cite{tusimple}, LLAMAS\cite{llamas2019}. 

\textbf{CULane:}  CULane is a challenging dataset that contains various scenes: normal, crowded, dazzled light, shadow, no line, arrow, curve, and night. Its environments consist of urban and highway. It consists of 88880 training images and 34680 testing images. Its image resolution is \(590 \times  1640\).  

\textbf{Tusimple:} Tusimple is a dataset in highway scenes. It contains 3268 images for training, 358 images for validation, and 2782 images for testing. Its image resolution is \(720 \times 1280\).

\textbf{LLAMAS:} LLAMAS is a highway lane detection dataset too, it contains over 100k images, 58269 images for training, 20844 images for validation, and 20929 images for testing. Its image resolution is \(717 \times  1276\). Its test set's label isn't public, and the testing result will be given after uploading results to their website.

\subsection{Ecaluation Metrics}
For CULane and LLAMAS, we use F1-measure as the evaluation metric. If the Intersection-over-union(IoU) between predictions and ground truth is larger than 0.5, the predictions are considered as true positive. F1-measure is formulated as follows: 
\begin{equation}
\mathrm{F} 1=\frac{2 \times  Precision  \times   Recall }{ Precision + Recall },
\label{eq:F1}
\end{equation}
where $ Precision=\frac{TP}{TP+FP} $, $ Recall=\frac{TP}{TP+FN} $, $ FP $ is false positive and $ FN $ is false negative. 

For Tusimple, the official evaluation metric is accuracy, which is calculated as follows:
\begin{equation}
	\text { accuracy }=\frac{\sum_{c l i p} C_{c l i p}}{\sum_{c l i p} S_{c l i p}},
	\label{eq:accuracy}
\end{equation}
where $ C_{c l i p} $ is points which are predicted rightly, $ S_{c l i p} $ is the number of ground truth points. We also calculate the F1 score for Tusimple.

\begin{table*}[tp]
	\centering
	\resizebox{\linewidth}{!}{
		\begin{tabular}{llllcc ccccc ccccc cc}  
			\toprule
			
			& &	& & &	& &	&  CULane & &	& &	& & & Tusimple & &	\\
			\cmidrule(r){2-14} \cmidrule(r){15-18}
			\textbf{Method} &  & \textbf{FPS} & \textbf{Device}  & \textbf{Total} & \textbf{Normal}& \textbf{Crowd}& \textbf{Dazzle}& \textbf{Shadow}  &\textbf{No line}& \textbf{Arrow}& \textbf{Curve}& \textbf{Cross}& \textbf{ Night}  &\textbf{F1\%}  &  \textbf{ Acc\%} &\textbf{FP\%} &\textbf{FN\%} \\		
			\midrule
			\textbf{Segmentation-based}& \\
			\cmidrule(r){1-2}
			SCNN(VGG16)\cite{pan2018spatial}    &  &7.5& GTX Titan Black & 71.60 &90.60 & 69.70 &58.50 & 66.90& 43.40&84.10&64.40&1990&66.10&95.97 &96.53&6.17&\textbf{1.80} \\
			RESA(ResNet-34)\cite{zheng2021resa} &  &\textbf{45.5}&  RTX 2080Ti & 74.50 &91.90 & 72.40 &66.50 & 72.00& 46.30&88.10&68.60&1896&69.80&\textbf{96.93} &\textbf{96.82}&\textbf{3.63}&2.84 \\
			RESA(ResNet-50)\cite{zheng2021resa} &  &35.7& RTX 2080Ti  &  75.30 &\textbf{92.10} & 73.10 &69.20 & 72.80& 47.70&\textbf{88.30}&70.30&1503&69.90&- &-&-&- \\
			LaneAF(ERFNet)\cite{abualsaud2021laneaf} &   & 24  & GTX 1080Ti & 75.63 &91.10 & 73.32 &69.71 & 75.81& 50.62&86.86&65.02&1844&70.90&- &-&-&- \\
			LaneAF(DLA-34)\cite{abualsaud2021laneaf} &   & 20  & GTX 1080Ti   & \textbf{77.41} &91.80 & \textbf{75.61} &\textbf{71.78 }& \textbf{79.12}& \textbf{51.38}&86.88&\textbf{72.70}&\textbf{1360}&\textbf{73.03}&- &-&-&-  \\
			\midrule
			\textbf{Line-anchor-based}& \\
			\cmidrule(r){1-2}
			LaneATT(ResNet-18)\cite{tabelini2021keep} &&\textbf{250} & RTX 2080Ti  & 75.13 &91.17 & 72.71 &65.82 & 68.03& 49.13&87.82&63.75&1020&68.58&96.71 &95.57&3.56&3.01 \\
			LaneATT(ResNet-34)\cite{tabelini2021keep} &&171 & RTX 2080Ti & 76.68 &92.14 & 75.03 &66.47 & 78.15& 49.39&88.38&67.72&1330&70.72&96.77 &95.63&3.53&2.92 \\
			LaneATT(ResNet-122)\cite{tabelini2021keep} &&26 & RTX 2080Ti  & 77.02 &91.74 & 76.16 &69.47 & 76.31& 50.46&86.29&64.05&1264&70.81&96.06&96.10&5.64&2.17 \\
			CLRNet(ResNet-18)\cite{zheng2022clrnet} &&119	& GTX 1080Ti & 79.58 &93.30 & 78.33 &73.71 & 79.66& 53.14&90.25&71.56&1321&75.11&\textbf{97.89} &96.84&2.28&\textbf{1.92} \\
			CLRNet(ResNet-34)\cite{zheng2022clrnet} &&103 & GTX 1080Ti  & 79.73 &93.49 & 78.06 &74.57 & 79.92& 54.01&90.59&72.77&1216&75.02&97.82 &\textbf{96.87}&\textbf{2.27}&2.08  \\
			CLRNet(ResNet-101)\cite{zheng2022clrnet} &&46  & GTX 1080Ti & 80.13 &\textbf{93.85} & 78.78 &72.49 & 82.33& 54.50&89.79&\textbf{75.57}&1262&\textbf{75.51}&97.62 &96.83&2.37&2.38  \\
			CLRNet(DLA-34)\cite{zheng2022clrnet} &&94  & GTX 1080Ti  & \textbf{80.47} &93.73 & \textbf{79.59} &\textbf{75.30} & \textbf{82.51}& \textbf{54.58}&\textbf{90.62}&74.13&\textbf{1155}&75.37&- &-&-&-  \\
			\midrule
			\textbf{Row-anchor-based}& \\
			\cmidrule(r){1-2}
			UFLD(ResNet-18)\cite{qin2020ultra} &&323 & GTX 1080Ti & 68.40 &87.70 & 66.00 &58.40 & 62.80& 40.20&81.00&57.90&1743&62.10&87.87 &95.82&19.05&3.92 \\
			UFLD(ResNet-34)\cite{qin2020ultra} &&175 &GTX 1080Ti  & 72.30 &90.70 & 70.20 &59.50 & 69.30& 44.40&85.70&69.50&2037&66.70&88.02 &95.86&18.91&3.75 \\
			UFLDv2(ResNet-18)\cite{qin2022ultra} &&\textbf{330} & RTX 3090  & 74.7 &91.7 & 73.0 &64.6 & 74.7& 47.2&87.6&68.7&1998&70.2&96.16&95.65&3.06&4.61 \\
			UFLDv2(ResNet-34)\cite{qin2022ultra} &&165 & RTX 3090	& 75.9 &92.5 & 74.9 &65.7 & 75.3& 49.0&88.5&70.2&1864&70.6&96.22 &95.56&3.18&4.37 \\
			CondLaneNet(ResNet-18)\cite{liu2021condlanenet} &&220 & RTX 2080  & 78.14 &92.87 & 75.79 &70.72 & 80.01& 52.39&89.37&72.40&1364&73.23&97.01 &95.84&2.18&3.80  \\
			CondLaneNet(ResNet-34)\cite{liu2021condlanenet} &&152 &  RTX 2080   & 78.74 &93.38 & 77.14 &\textbf{71.17} & 79.93& 51.85&89.89&73.88&1387&73.92&96.98 &95.37&2.20&3.82  \\
			CondLaneNet(ResNet-101)\cite{liu2021condlanenet} &&58 &  RTX 2080   & \textbf{79.48} &\textbf{93.47} & \textbf{77.44} &70.93 & \textbf{80.91}& \textbf{54.13}&\textbf{90.16}&\textbf{75.21}&\textbf{1201}&\textbf{74.80}&\textbf{97.24} &\textbf{96.54}&\textbf{2.01}&\textbf{3.50}  \\
			\midrule
			\textbf{Keypoint-based}& \\
			\cmidrule(r){1-2}
			FOLOLane(ERFNet)\cite{qu2021focus}   &&40 & Tesla-V100 & 78.80 &92.70 & 77.80 &75.20 &79.30& 52.10& 89.00&69.40&1569&\textbf{74.50}&96.59 &\textbf{96.92}&4.47&2.28 \\
			GANet-S(ResNet-18)\cite{wang2022keypoint} &&\textbf{153} & Tesla-V100& 78.79 &93.24 & 77.16 &71.24 & 77.88& \textbf{53.59}&89.62&75.92&\textbf{1240}&72.75&\textbf{97.71}&95.95&\textbf{1.97}&2.62 \\
			GANet-M(ResNet-34)\cite{wang2022keypoint} &&127 & Tesla-V100& 79.39 &\textbf{93.73} & 77.92 &71.64 & \textbf{79.49}& 52.63&\textbf{90.37}&76.32&1368&73.67&97.68 &95.87&1.99&2.64 \\
			GANet-L(ResNet-101)\cite{wang2022keypoint} &&63 & Tesla-V100	& \textbf{79.63} &93.67 & \textbf{78.66} &\textbf{71.82} & 78.32& 53.38&89.86&\textbf{77.37}&1352&73.85&97.45 &96.44&2.63&\textbf{2.47} \\

			\midrule
			\textbf{Curve-based}& \\
			\cmidrule(r){1-2}
			PolyLaneNet(ResNet-18)\cite{tabelini2021polylanenet} &&- &- &- & -&-& -&- & -&- &- & -&- &90.62 &93.36&9.42&9.33 \\
			
			LaneFormer(ResNet-18)\cite{han2022laneformer}  &&- &- &71.71&88.60 & 69.02 & 64.07& 65.02& 45.00& 81.55& 60.46& 25& 64.76&-&96.54 &4.35& 2.36 \\
			LaneFormer(ResNet-34)\cite{han2022laneformer}  &&-&- & 74.70&90.74 &72.31 &69.12 &71.57 &47.37 &85.07 &65.90  &26 &67.77& - &96.56 &5.39 &3.37 \\
			LaneFormer(ResNet-50)\cite{han2022laneformer}  &&53 & Tesla-V100&77.06&91.77&75.41 &70.17 &75.75 &48.73& 87.65 &66.33  &\textbf{19} &71.04 & - &96.80 &5.60 &\textbf{1.99} \\
			BézierLaneNet(ResNet-18)\cite{feng2022rethinking}   &&\textbf{213} & RTX 2080Ti	& 73.67 &90.22 & 71.55 &62.49 & 70.91& 45.30&84.09&58.98&996&68.70&- &95.41&5.3&4.6 \\
			BézierLaneNet(ResNet-34)\cite{feng2022rethinking}   &&150 &RTX 2080Ti	& 75.57  &91.59  &73.20 &69.20 & 76.74&48.05&87.16 &62.45 &888&69.90 &-&95.65&5.1&3.9 \\
			\textbf{BSNet(ResNet-18)}    &&197 & GTX 1080Ti &79.64&93.46&77.93&74.25&81.95&54.24&90.05&73.62&1400&75.11      & {97.79}& {96.63}& {2.03}& {2.39}\\
			\textbf{BSNet(ResNet-34)}    && {133} & GTX 1080Ti & {79.89}& {93.75}& {78.01}& \textbf{76.65}& {79.55}& {54.69}& {90.72}& {73.99}& {1445}& {75.28 }   & {97.79}& \textbf{96.78}& {2.12}& {2.32}\\
			\textbf{BSNet(ResNet-101)}   &  &  {48} & GTX 1080Ti& {80.00} & {93.75}& {78.44}& {74.07}&  {81.51}& {54.83}& {90.48}& {74.01}& {1255}& {75.12}& \textbf{97.84}& {96.73}& \textbf{1.98}& {2.35} \\
			\textbf{BSNet(DLA-34)}    & & {119} & GTX 1080Ti & \textbf{80.28}& \textbf{93.87}& \textbf{78.92}& {75.02}& \textbf{82.52}& \textbf{54.84}& \textbf{90.73}& \textbf{74.71}& {1485}& \textbf{75.59}& {97.70}& {96.70} & {2.12}& {2.48}  \\
			\bottomrule
		\end{tabular}
	} 
	\caption{Results on test set of CULane and Tusimple. }
	\label{tab:culane_tusimple}
\end{table*}

\subsection{Implementation Details}
We adopt ResNet\cite{he2016deep} and DLA\cite{yu2018deep} as our pre-trained backbones. All input images of all datasets are resized to \(320\times 800\), which means $ H=320, W=800 $. We followed data augmentation used in \cite{liu2021condlanenet,qu2021focus,zheng2022clrnet}, which are random affine transformation (translation, rotation, and scaling), and random horizontal flips. In training sessions, we use AdamW\cite{IlyaLoshchilov2017DecoupledWD} optimizer to train three datasets, and the learning rate is initialized as 1e-3 and with cosine decay learning rate strategy\cite{IlyaLoshchilov2016SGDRSG} with power set to 0.9. We set the number of proposals \(N_p = 60\), proposals' feature dimension $C=256$, lanes' extended radius $r=9$, the number of control points is 8, the number of sample points for distance calculation \(N_{dis}=300\), the number of sample points for local features $N_{fea}=30$. The training numbers of epochs for  CULane, Tusimple, and LLAMAS are 15, 100, and 20. All experiments are implemented based on Pytorch\cite{paszke2019pytorch} with 1 GPU.

\begin{table}[tp]
	\centering
	\resizebox{\linewidth}{!}{
		\begin{tabular}{lllccc}  
			\toprule
			 & \multicolumn{4}{c}{\textbf{ LLAMAS}} \\
	
			\cmidrule(r){2-6}
			\textbf{Method}&\textbf{FPS}&\textbf{Device} &\textbf{F1}&\textbf{Precision}&\textbf{Recall} \\ 
			\midrule

			\textbf{Line-anchor-based}\\
			\cmidrule(r){1-1}
			LaneATT(ResNet-18)\cite{tabelini2021keep}   & \textbf{250} & RTX 2080Ti & 93.46 & 96.92 & 90.24 \\
			LaneATT(ResNet-34)\cite{tabelini2021keep}  &  171 & RTX 2080Ti  & 93.74 & 96.79 & \textbf{90.88} \\
			LaneATT(ResNet-122)\cite{tabelini2021keep} & 26  & RTX 2080Ti& 93.54 & \textbf{96.82} & 90.47 \\
			CLRNet(ResNet-18)\cite{zheng2022clrnet}  & 119  & GTX 1080Ti& 96.00 & - & - \\
			CLRNet(DLA34)\cite{zheng2022clrnet}   & 94 & GTX 1080Ti & \textbf{96.12} &- & - \\
			\midrule
			\textbf{Curve-based}\\
			\cmidrule(r){1-1}
			PolyLaneNet(EfficientNet-B0)\cite{tabelini2021polylanenet} &- &- &88.40 &88.87 &87.93 \\
			BézierLaneNet(ResNet-18)\cite{feng2022rethinking}  &\textbf{213}  & RTX 2080Ti  & 94.91 & 95.71 & 94.13 \\
			BézierLaneNet(ResNet-34)\cite{feng2022rethinking}  &150   & RTX 2080Ti  & 95.17 & 95.89 & 94.46 \\
			\textbf{BSNet(ResNet-18)}   &197& GTX 1080Ti &95.97  & 96.80 & 95.16 \\
			\textbf{BSNet(ResNet-34)}  & 133& GTX 1080Ti  &96.13 & 96.92 & 95.35 \\	
			\textbf{BSNet(DLA-34)}   & 119 & GTX 1080Ti& \textbf{96.15} & \textbf{96.96} &\textbf{ 95.36} \\	
			\bottomrule
		\end{tabular}
	}
	\caption{Results on test set of LLAMAS.  }
	\label{tab:llamas}
\end{table}

\subsection{Quantiative Results}
\textbf{Performance on CULane}. We show the test results on CULane in \cref{tab:culane_tusimple}. Our method achieves state-of-the-art results on CULane and obtains the same level of F1-measure with CLRNet\cite{zheng2022clrnet} when using the same backbone, and our method is much faster, our ResNet-18 version model reaches 197 FPS on NVIDIA 1080Ti GPU, it is 75+ FPS faster than ResNet-18 version CLRNet, our ResNet-34 version model even faster than the ResNet-18 version CLRNet. Our method greatly improves accuracy in nearly all scenes among curve-based methods, especially improves the performance on curvy scenes.

\textbf{Performance on Tusimple}  As shown in \cref{tab:culane_tusimple}, the performances of different methods are very close because the accuracy of the dataset is nearly statured now. Our method achieves high F1, high accuracy, low FP, and low FN simultaneously, performs much better than past curve-based methods, and has the same level of F1-measure with other state-of-the-art methods (CLRNet\cite{zheng2022clrnet}, GANet\cite{wang2022keypoint}), but our models' speed is much faster.

\textbf{Performance on LLAMAS} Our methods get state-of-the-art results on LLAMAS, our DLA-34 version BSNet achieves 96.15\% F1-measure and 119 FPS. Although the ResNet-18 version has a little lower F1, but its speed is much faster.

Our methods prove that curve-based methods are not only an intuitive and concise representation of lanes but also have the ability to achieve high accuracy and fast speed simultaneously. The results above show our method is very competitive. Its prediction results are holistic curves, and b-spline curves can fit lanes flexibly, even in curvy scenes. Our simple but efficient network can provide enough global features and local features for curve fitting. What's more, our networks have fast convergence speed. For example, when training Tusimple, our ResNet-18 version method's accuracy can reach 95.73\% after about 20 epochs, which is better than the final result that trained 400 epochs of BézierLaneNet\cite{feng2022rethinking}  This means our loss function is very fitted with the lane detection optimization goal, and the ill-condition problems are alleviated.

Our methods use 60 proposals that are more sparse than line anchors in line anchor methods and don't rely on complex refinements in CLRNet\cite{zheng2022clrnet}. BSNet is very simple, like LaneATT\cite{tabelini2021keep}, but it is more flexible as don't rely on prior information; proposals are predicted with global features and can easily fit horizon curves which are hard for line anchors and row anchors to fit.

\subsection{Ablation study}

We conducted extensive experiments on the CULane dataset to verify our methods. All experiments of the ablation study are based on ResNet-18 version BSNet. The results are shown in \cref{tab:ablation}. The experiments are set from four aspects: Lane Representation, Proposals' Features, Regression Loss, and Local feature and attention mechanism (LA). 
Lane representations contain bézier curve and b-spline curve. Proposals' features can be obtained from the column pooling module in BézierLaneNet or our proposals' features init module (PFIM). Regression losses can be calculated as the original regression loss in BézierLaneNet\cite{feng2022rethinking} and our regression loss ($ L_{reg} $,$ L_{length} $,$ L_{start} $ are used), and these two regression losses calculate methods are written as $ original $ and $ new $ in  \cref{tab:ablation}.

\textbf{Ablation study on PFIM.} We can see results in the first row and second row in \cref{tab:ablation}. The F1 increased from 74.47 to 77.64, which proves the global feature obtained from PFIM is very efficient. And the number of proposals is limited to image resolution when using column pooling in BézierLanNet\cite{feng2022rethinking}, but PFIM can easily change the number of proposals by changing the number of image feature channels. It is useful to adjust the number of proposals to cope with different road scenes.

\textbf{Ablation study on lane representation.} We can see results in the second row and third row in \cref{tab:ablation}. The F1 increased from 77.64 to 78.14, which shows that b-spline curves have stronger fitting ability than bézier curves. It's worth noticing that the F1 difference between the second row and third row is bigger than the difference between the fifth row and sixth row because the local feature in LA helps curves to refine, which makes up for the lack of fitting ability of bézier curves partially.

\textbf{Ablation study on regression loss.} We can see results in the third row and fourth row in \cref{tab:ablation}. The F1 increased from 78.14 to 78.88, which shows that our new regression loss is more reasonable for lane line fitting.

\textbf{Ablation study on local feature and attention mechanism}. We can see results in the fourth row and sixth row in \cref{tab:ablation}. The F1 increased from 78.88 to 79.64, which shows that the local features and relationships between proposals are useful for locating precise lane positions. 

\textbf{Ablation study on the number of control points} We also conducted some experiments to know how many control points should be used. We can see the results in \cref{tab:number} , when we use eight control points, the F1 gets maximum. In fact, the number of control points depends on the status of lanes in datasets. If the lanes' topology is very simple, even the bézier curve can get well results, but if the lanes' topology is complex, we need to increase the number of control points. And the number of control points can't be too much, or it will decrease the curve's globality. The globality and locality should keep balance.

\begin{table}[tp]
	\centering
	\resizebox{\linewidth}{!}{
		\begin{tabular}{cccccccc}  
			\toprule
			\multicolumn{2}{c}{\textbf{Lane representation}} &\multicolumn{2}{c}{\textbf{Proposals' Features}} & \multicolumn{2}{c}{\textbf{Regression Loss}}& \\
			
			\cmidrule(r){1-2} \cmidrule(r){3-4} \cmidrule(r){5-6}
			
			Bézier &B-spline  & Column pooling& PFIM  & origin& new & LA &\textbf{F1}   \\
			\midrule
			\checkmark&       &\checkmark&         &\checkmark&             && 74.47 \\
			\checkmark   &&&\checkmark         &\checkmark&             & &77.64 \\
			&\checkmark&    &\checkmark&\checkmark&             & &78.14 \\
			&\checkmark&&\checkmark &&\checkmark  & &78.88\\
			\checkmark&&&\checkmark &&\checkmark  & \checkmark &79.32 \\
			&\checkmark&&\checkmark &&\checkmark  & \checkmark &79.64 \\
			
			\bottomrule

		\end{tabular}
	}
	
	\caption{Effects of each component in our method. Results are reported on CULane. }
	\label{tab:ablation}
\end{table}

\begin{table}[tp]
	\centering
	\resizebox{1\linewidth}{!}{
		\begin{tabular}{lcccccccccc}  
			\toprule
			 &  \textbf{Number}& 4&6 &8&10&12&14&16&18\\
			 	\midrule
			  & \textbf{F1}& 79.32&79.39&79.64&79.57&79.61&79.59&79.47&79.53 \\
		
			\bottomrule

		\end{tabular}
	}
	
	\caption{Results of ablation study on the number of control points. Results are reported on CULane.}
	\label{tab:number}
\end{table}

\section{Conclusion}

We propose a deep lane detector BSNet based on the b-spline curve.
Our proposed new curve regression loss alleviates the ill-conditioned problems and makes the optimization objective more reasonable. 
The b-spline curve can easily represent curvy lanes and horizon lanes that are hard for some other lane representations to fit.
Our experiments show that the curve-based method can also have a strong fitting ability, and BSNet can provide enough global and local features with a simple structure.
Our methods achieve state-of-the-art results on three widely used datasets. 
BSNet performs high accuracy and fast speed simultaneously.  
We hope our methods can inspire more work using parametric curves to solve lane detection tasks elegantly. It can make the lane detection task more like the usual object detection task\cite{ren2015faster,sun2021sparse}, except representations are different, which are bounding box and b-spline curves. Then these two tasks' networks can be joined into one network easier. And our network can perhaps be used in object contour detection\cite{yang2016object} for precise location. The strong fitting ability of BSNet can release big power among relative fields.

{\small
	\bibliographystyle{ieee_fullname}
	\bibliography{egbib.bib}

\begin{thebibliography}{10}\itemsep=-1pt

\bibitem{abualsaud2021laneaf}
Hala Abualsaud, Sean Liu, David~B Lu, Kenny Situ, Akshay Rangesh, and Mohan~M
  Trivedi.
\newblock Laneaf: Robust multi-lane detection with affinity fields.
\newblock {\em IEEE Robotics and Automation Letters}, 6(4):7477--7484, 2021.

\bibitem{llamas2019}
Karsten Behrendt and Ryan Soussan.
\newblock Unsupervised labeled lane markers using maps.
\newblock In {\em Proceedings of the IEEE International Conference on Computer
  Vision}, 2019.

\bibitem{bolya2019yolact}
Daniel Bolya, Chong Zhou, Fanyi Xiao, and Yong~Jae Lee.
\newblock Yolact: Real-time instance segmentation.
\newblock In {\em Proceedings of the IEEE/CVF international conference on
  computer vision}, pages 9157--9166, 2019.

\bibitem{carion2020end}
Nicolas Carion, Francisco Massa, Gabriel Synnaeve, Nicolas Usunier, Alexander
  Kirillov, and Sergey Zagoruyko.
\newblock End-to-end object detection with transformers.
\newblock In {\em European conference on computer vision}, pages 213--229.
  Springer, 2020.

\bibitem{feng2022rethinking}
Zhengyang Feng, Shaohua Guo, Xin Tan, Ke Xu, Min Wang, and Lizhuang Ma.
\newblock Rethinking efficient lane detection via curve modeling.
\newblock In {\em Proceedings of the IEEE/CVF Conference on Computer Vision and
  Pattern Recognition}, pages 17062--17070, 2022.

\bibitem{girshick2015fast}
Ross Girshick.
\newblock Fast r-cnn.
\newblock In {\em Proceedings of the IEEE international conference on computer
  vision}, pages 1440--1448, 2015.

\bibitem{han2022laneformer}
Jianhua Han, Xiajun Deng, Xinyue Cai, Zhen Yang, Hang Xu, Chunjing Xu, and
  Xiaodan Liang.
\newblock Laneformer: Object-aware row-column transformers for lane detection.
\newblock {\em arXiv preprint arXiv:2203.09830}, 2022.

\bibitem{he2016deep}
Kaiming He, Xiangyu Zhang, Shaoqing Ren, and Jian Sun.
\newblock Deep residual learning for image recognition.
\newblock In {\em Proceedings of the IEEE conference on computer vision and
  pattern recognition}, pages 770--778, 2016.

\bibitem{ko2021key}
Yeongmin Ko, Younkwan Lee, Shoaib Azam, Farzeen Munir, Moongu Jeon, and Witold
  Pedrycz.
\newblock Key points estimation and point instance segmentation approach for
  lane detection.
\newblock {\em IEEE Transactions on Intelligent Transportation Systems}, 2021.

\bibitem{li2019line}
Xiang Li, Jun Li, Xiaolin Hu, and Jian Yang.
\newblock Line-cnn: End-to-end traffic line detection with line proposal unit.
\newblock {\em IEEE Transactions on Intelligent Transportation Systems},
  21(1):248--258, 2019.

\bibitem{lin2017feature}
Tsung-Yi Lin, Piotr Doll{\'a}r, Ross Girshick, Kaiming He, Bharath Hariharan,
  and Serge Belongie.
\newblock Feature pyramid networks for object detection.
\newblock In {\em Proceedings of the IEEE conference on computer vision and
  pattern recognition}, pages 2117--2125, 2017.

\bibitem{lin2017focal}
Tsung-Yi Lin, Priya Goyal, Ross Girshick, Kaiming He, and Piotr Doll{\'a}r.
\newblock Focal loss for dense object detection.
\newblock In {\em Proceedings of the IEEE international conference on computer
  vision}, pages 2980--2988, 2017.

\bibitem{liu2021condlanenet}
Lizhe Liu, Xiaohao Chen, Siyu Zhu, and Ping Tan.
\newblock Condlanenet: a top-to-down lane detection framework based on
  conditional convolution.
\newblock In {\em Proceedings of the IEEE/CVF International Conference on
  Computer Vision}, pages 3773--3782, 2021.

\bibitem{liu2021end}
Ruijin Liu, Zejian Yuan, Tie Liu, and Zhiliang Xiong.
\newblock End-to-end lane shape prediction with transformers.
\newblock In {\em Proceedings of the IEEE/CVF winter conference on applications
  of computer vision}, pages 3694--3702, 2021.

\bibitem{IlyaLoshchilov2016SGDRSG}
Ilya Loshchilov and Frank Hutter.
\newblock Sgdr: Stochastic gradient descent with warm restarts.
\newblock {\em arXiv: Learning}, 2016.

\bibitem{IlyaLoshchilov2017DecoupledWD}
Ilya Loshchilov and Frank Hutter.
\newblock Decoupled weight decay regularization.
\newblock {\em Learning}, 2017.

\bibitem{neven2018towards}
Davy Neven, Bert De~Brabandere, Stamatios Georgoulis, Marc Proesmans, and Luc
  Van~Gool.
\newblock Towards end-to-end lane detection: an instance segmentation approach.
\newblock In {\em 2018 IEEE intelligent vehicles symposium (IV)}, pages
  286--291. IEEE, 2018.

\bibitem{pan2018spatial}
Xingang Pan, Jianping Shi, Ping Luo, Xiaogang Wang, and Xiaoou Tang.
\newblock Spatial as deep: Spatial cnn for traffic scene understanding.
\newblock In {\em Proceedings of the AAAI Conference on Artificial
  Intelligence}, volume~32, 2018.

\bibitem{paszke2019pytorch}
Adam Paszke, Sam Gross, Francisco Massa, Adam Lerer, James Bradbury, Gregory
  Chanan, Trevor Killeen, Zeming Lin, Natalia Gimelshein, Luca Antiga, et~al.
\newblock Pytorch: An imperative style, high-performance deep learning library.
\newblock {\em Advances in neural information processing systems}, 32, 2019.

\bibitem{qin2020ultra}
Zequn Qin, Huanyu Wang, and Xi Li.
\newblock Ultra fast structure-aware deep lane detection.
\newblock In {\em European Conference on Computer Vision}, pages 276--291.
  Springer, 2020.

\bibitem{qin2022ultra}
Zequn Qin, Pengyi Zhang, and Xi Li.
\newblock Ultra fast deep lane detection with hybrid anchor driven ordinal
  classification.
\newblock {\em IEEE Transactions on Pattern Analysis and Machine Intelligence},
  2022.

\bibitem{qu2021focus}
Zhan Qu, Huan Jin, Yang Zhou, Zhen Yang, and Wei Zhang.
\newblock Focus on local: Detecting lane marker from bottom up via key point.
\newblock In {\em Proceedings of the IEEE/CVF Conference on Computer Vision and
  Pattern Recognition}, pages 14122--14130, 2021.

\bibitem{ren2015faster}
Shaoqing Ren, Kaiming He, Ross Girshick, and Jian Sun.
\newblock Faster r-cnn: Towards real-time object detection with region proposal
  networks.
\newblock {\em Advances in neural information processing systems}, 28, 2015.

\bibitem{su2021structure}
Jinming Su, Chao Chen, Ke Zhang, Junfeng Luo, Xiaoming Wei, and Xiaolin Wei.
\newblock Structure guided lane detection.
\newblock {\em arXiv preprint arXiv:2105.05403}, 2021.

\bibitem{sun2021sparse}
Peize Sun, Rufeng Zhang, Yi Jiang, Tao Kong, Chenfeng Xu, Wei Zhan, Masayoshi
  Tomizuka, Lei Li, Zehuan Yuan, Changhu Wang, et~al.
\newblock Sparse r-cnn: End-to-end object detection with learnable proposals.
\newblock In {\em Proceedings of the IEEE/CVF conference on computer vision and
  pattern recognition}, pages 14454--14463, 2021.

\bibitem{tabelini2021keep}
Lucas Tabelini, Rodrigo Berriel, Thiago~M Paixao, Claudine Badue, Alberto~F
  De~Souza, and Thiago Oliveira-Santos.
\newblock Keep your eyes on the lane: Real-time attention-guided lane
  detection.
\newblock In {\em Proceedings of the IEEE/CVF conference on computer vision and
  pattern recognition}, pages 294--302, 2021.

\bibitem{tabelini2021polylanenet}
Lucas Tabelini, Rodrigo Berriel, Thiago~M Paixao, Claudine Badue, Alberto~F
  De~Souza, and Thiago Oliveira-Santos.
\newblock Polylanenet: Lane estimation via deep polynomial regression.
\newblock In {\em 2020 25th International Conference on Pattern Recognition
  (ICPR)}, pages 6150--6156. IEEE, 2021.

\bibitem{tusimple}
TuSimple.
\newblock Tusimple benchmark.
\newblock \url{https://github.com/TuSimple/tusimple-benchmark/}, Accessed
  September 2020.

\bibitem{wang2022keypoint}
Jinsheng Wang, Yinchao Ma, Shaofei Huang, Tianrui Hui, Fei Wang, Chen Qian, and
  Tianzhu Zhang.
\newblock A keypoint-based global association network for lane detection.
\newblock In {\em Proceedings of the IEEE/CVF Conference on Computer Vision and
  Pattern Recognition}, pages 1392--1401, 2022.

\bibitem{xu2020curvelane}
Hang Xu, Shaoju Wang, Xinyue Cai, Wei Zhang, Xiaodan Liang, and Zhenguo Li.
\newblock Curvelane-nas: Unifying lane-sensitive architecture search and
  adaptive point blending.
\newblock In {\em European Conference on Computer Vision}, pages 689--704.
  Springer, 2020.

\bibitem{yang2016object}
Jimei Yang, Brian Price, Scott Cohen, Honglak Lee, and Ming-Hsuan Yang.
\newblock Object contour detection with a fully convolutional encoder-decoder
  network.
\newblock In {\em Proceedings of the IEEE conference on computer vision and
  pattern recognition}, pages 193--202, 2016.

\bibitem{yu2018deep}
Fisher Yu, Dequan Wang, Evan Shelhamer, and Trevor Darrell.
\newblock Deep layer aggregation.
\newblock In {\em Proceedings of the IEEE conference on computer vision and
  pattern recognition}, pages 2403--2412, 2018.

\bibitem{zheng2021resa}
Tu Zheng, Hao Fang, Yi Zhang, Wenjian Tang, Zheng Yang, Haifeng Liu, and Deng
  Cai.
\newblock Resa: Recurrent feature-shift aggregator for lane detection.
\newblock In {\em Proceedings of the AAAI Conference on Artificial
  Intelligence}, volume~35, pages 3547--3554, 2021.

\bibitem{zheng2022clrnet}
Tu Zheng, Yifei Huang, Yang Liu, Wenjian Tang, Zheng Yang, Deng Cai, and
  Xiaofei He.
\newblock Clrnet: Cross layer refinement network for lane detection.
\newblock In {\em Proceedings of the IEEE/CVF Conference on Computer Vision and
  Pattern Recognition}, pages 898--907, 2022.

\end{thebibliography}
}

\end{document}